\documentclass[sigconf]{acmart}

\renewcommand\footnotetextcopyrightpermission[1]{} 
\AtBeginDocument{%
  \providecommand\BibTeX{{%
    \normalfont B\kern-0.5em{\scshape i\kern-0.25em b}\kern-0.8em\TeX}}
    }

\usepackage{amsmath,amsfonts}
\usepackage{algorithmic}
\usepackage{braket}
\usepackage{textcomp}
\usepackage{xcolor}
\usepackage{hyperref}
\usepackage{bm}
\usepackage{booktabs}
\usepackage{epstopdf} 
\usepackage{multirow}
\usepackage{subfigure}
\usepackage{xspace}
\usepackage{url}
\usepackage{graphicx}
\usepackage{float}
\usepackage{mathtools, nccmath}

\def\Ex{\textbf{$\mathcal E$}}

\makeatletter
\newcommand{\thickhline}{%
    \noalign {\ifnum 0=`}\fi \hrule height 0.65pt
    \futurelet \reserved@a \@xhline
}
\makeatother

\newcommand{\COMMENT}{\textcolor{red}}

\begin{document}

\title[Physics-Guided Problem Decomposition for Scaling Deep Learning of High-dimensional Eigen-Solvers]{Physics-Guided Problem Decomposition for Scaling \\
Deep Learning of High-dimensional Eigen-Solvers: \\ 
The Case of Schr\"{o}dinger's Equation}

\author{Sangeeta Srivastava}
\affiliation{%
  \institution{The Ohio State University}}

\author{Samuel Olin}
\affiliation{%
  \institution{Binghamton University}}

\author{Viktor Podolskiy}
\affiliation{%
  \institution{University of Massachusetts Lowell}}

\author{Anuj Karpatne}
\affiliation{%
  \institution{Virginia Tech}}

\author{Wei-Cheng Lee}
\affiliation{%
  \institution{Binghamton University}}

\author{Anish Arora}
\affiliation{%
  \institution{The Ohio State University}}

\begin{abstract}
Given their ability to effectively learn non-linear mappings and perform fast inference, deep neural networks (NNs) have been proposed as a viable alternative to traditional simulation-driven approaches for solving high-dimensional eigenvalue equations (HDEs), which are the foundation for many scientific applications. Unfortunately, for the learned models in these scientific applications to achieve generalization, a large, diverse, and preferably annotated dataset is typically needed and is computationally expensive to obtain. Furthermore, the learned models tend to be memory-- and compute--intensive primarily due to the size of the output layer. While generalization, especially extrapolation, with scarce data has been attempted by imposing physical constraints in the form of physics loss, the problem of model scalability has remained. 

In this paper, we alleviate the compute bottleneck in the output layer by using physics knowledge to decompose the \textit{complex} regression task of predicting the high-dimensional eigenvectors into multiple \textit{simpler} sub-tasks, each of which are learned by a simple ``expert" network. We call the resulting architecture of  specialized experts \textit{Physics-Guided Mixture-of-Experts (PG-MoE)}. We demonstrate the efficacy of such physics-guided problem decomposition for the case of the Schr\"{o}dinger Equation in Quantum Mechanics. Our proposed PG-MoE model predicts the ground-state solution, i.e., the eigenvector that corresponds to the smallest possible eigenvalue.  The model is $150\times$ smaller than the network trained to learn the complex task while being competitive in generalization. To improve the generalization of the PG-MoE, we also employ a physics-guided loss function based on variational energy, which by quantum mechanics principles is minimized iff the output is the ground-state solution. 
\end{abstract}



\maketitle
\pagestyle{empty}

\section{Introduction}
While neural networks often achieve state-of-the-art performance and generalization in problems with large amounts of data, such as natural language processing (NLP) or computer vision, they often do not work as well in scientific applications where system behavior is governed by high-dimensional eigenvalue problems (HDEs).
This means that the power of neural networks is yet to be appropriately exploited in fields such as quantum mechanics, solid state physics, electromagnetism, and classical mechanics where HDEs are common. There are three main challenges that have limited the use of NNs in scientific fields: (\textbf{i}) Large amounts of annotated data are prohibitively expensive to obtain. (\textbf{ii}) Purely data-driven models may extrapolate to produce predictions on unseen data that are physically inconsistent, due to the bias induced by the training data. (\textbf{iii}) 
In many physical problems, the dimension of the eigenvectors scales in size with the number of system-dependent degrees of freedom, $f$; for a $N$-particle system, the dimension of the eigenvector  grows exponentially with $f$, i.e., $\mathcal O(f^N)$; and therefore, a neural network trained to predict an eigenvector shares a similar exponential growth in its output layer, making the learning intractable for high values of $N$.


We propose to leverage physics knowledge to overcome the aforementioned challenges of deep learning-based solutions for solving HDEs, yielding \textit{model scalability} and \textit{generalization}. For the former, i.e., to alleviate the memory requirements induced by exponential growth in the output layer, we first divide the given input-space into multiple input-subspaces in a physically sound manner and then train a specialized \textit{expert} network $\mathcal E$ for each input sub-space.  
We refer to the architecture comprising of all the \textit{experts} as \textit{\textbf{\underline{P}}hysics-\textbf{\underline{G}}uided \textbf{\underline{M}}ixture-\textbf{\underline{o}}f-\textbf{\underline{E}}xperts} \textbf{(PG-MoE)} (and describe it further in Section \ref{sec:arch}). The learning task of each \textit{expert} is ``simpler" in two ways: (1) each expert specializes in only its subset of the input space, and (2) instead of predicting all the $f^N$ coefficients of the eigenvector, each expert predicts only one coefficient at a time, allowing for independent learning of all the $f^N$ output values. This decomposition approach allows us to avoid the communication overhead among the experts, while also reducing the output space to a single coefficient of the eigenvector, both of which are critical to achieve scalability for large systems. For the latter, i.e., to address generalization, we add a physics-guided (PG) loss that penalizes the NN for violating  physical constraints, as has been done in previous work. By restricting the convergence space with PG loss, we are able to build physically consistent and scientifically sound predictive models even when there is scarcity of annotated data.

In this paper, we use Schr\"{o}dinger's equation in quantum mechanics as a case study to illustrate the benefits of our approach. We seek to find the eigenvector that corresponds to the smallest possible eigenvalue, also known as the \textit{ground-state} wavefunction. Since each particle in our problem possesses two quantum degrees of freedom - originating from its intrinsic angular momentum orientation - $\{\uparrow, \downarrow\}$ (described in Section \ref{sec:quantum}), the resulting eigenvector to be predicted has a dimension of $2^N$.
With the proposed combination of the physics-guided decomposition and physics-guided loss, we are able to produce a parameter-efficient, yet generalizable PG-MoE.\\

\noindent The contributions of this work are as follows:

\begin{enumerate}
    \item The proposed PG-MoE architecture trained on the sub-spaces is substantially smaller than the black-box baseline NN that is trained on the entire input space, and generalizes better on the unseen test data. As an example, for $N \! = \! 10$ and cosine similarity as the black-box loss, the PG-MoE NN is $\sim$150$\times$ smaller than the baseline NN. And to achieve similar test performance as the baseline, PG-MoE trains faster requiring $\sim$60$\%$ fewer epochs.

    \item 
    To better enforce generalization, we introduce a new PG loss function based on the variational method, which is well established in quantum mechanics theory \cite{Sakurai:1167961,Griffiths2004Introduction} for searching for the ground state wavefunction. The method is theoretically advantageous because it is specific to the ground state wavefunction, as opposed to the eigenvalue equation which is satisfied for any eigenvector. This ensures that the loss function possesses a global minimum if and only if the predicted wavefunction is the true ground state. Further, this loss does not require a computation of the ground state energy in either the NN or the input data.
    
    
    \item When physics loss is included in PG-MoE training, the cosine similarity on the unseen test set improves (by $\sim$11$\%$ for $N \! = \! 10$) compared to when only black-box loss is used. The similarity is also competitive to that of the baseline model trained with the PG loss.
    
    \item We show the impact of the design choices for PG-MoE training and identify the key parameters that must be tuned to enable convergence.
    
\end{enumerate}
\section{Quantum Mechanics}
\label{sec:quantum}

\begin{figure}[htb]
	\centering
	\includegraphics[width=\linewidth]{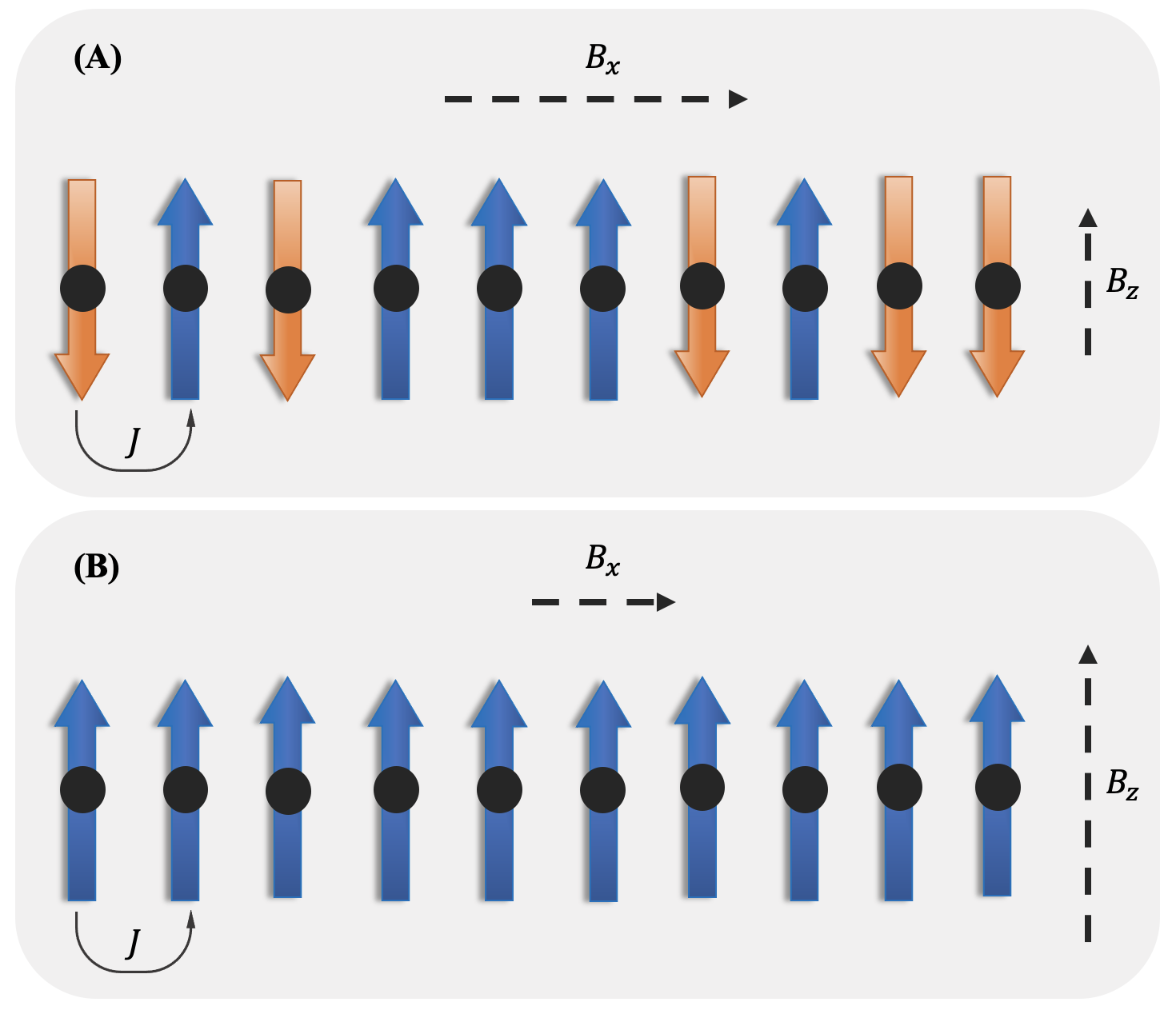}
	\caption{Ising spin chain for $N\!=\!10$ spins and coupling strength $J$. The black disks represent electrons at each site, and the arrows indicate the orientation of the spin. Box (A) shows a paramagnetic phase, with the net magnetization 1. Box (B) shows a ferromagnetic phase with net magnetization 5. The transverse field $B_x$, and longitudinal field $B_z$ are shown with vector length indicating field strength, which may be used to set the phase.}
	\label{fig:spin chain}
\end{figure}

An Ising model consists of $N$ interacting electrons, each of which possess an intrinsic degree of freedom called the ``spin" angular momentum. The possible values for the spin are defined with respect to an arbitrary axis, they are simply either the spin-up ($\uparrow$) or spin-down ($\downarrow$) state, which are typically expressed as a basis in Dirac notation: $\{ \ket{z; \uparrow}, \ket{z; \downarrow}\}$. Here, the z-axis is chosen, which is the convention for this work. A chain with $N$-spins can have $2^N$ unique combinations, also called \textit{configurations}. These configurations form an orthonormal basis for the Schr\"{o}dinger equation, resulting in a  ground state wavefunction with $2^N$ coefficients.

Here, we consider the one-dimensional transverse Ising model \cite{bonfim-ising} (Equation \ref{eq:1}). It describes a system of $N$ fixed spins and is useful to understand magnetism. The spins are subjected to both a transverse and longitudinal field, denoted $B_x$ and $B_z$ respectively. Neighboring spins are coupled through an interaction term, with strength $J$. Note that we fix $J=1$ for this work. The Schr\"{o}dinger eigenvalue equation, denoted as $\textbf{H}\Psi=E\Psi$, where \textbf{H} is the Hamiltonian (or "total energy") operator and $E$ is the energy eigenvalue, may be solved to obtain $\Psi$, the wavefunction, which carries all information about the system. The Hamiltonian for this system is
\begin{equation}\label{eq:1}
     \textbf{H}=-\frac{J}{N}\sum_{i=1}^N\sigma^z_i \sigma^z_{i+1}-\frac{1}{N}\sum_{i=1}^N \left( B_z \sigma^z_i + B_x \sigma^x_i \right)
\end{equation}
where $\sigma^{z,x}$ is the corresponding Pauli matrix, and the system is subject to a periodic boundary condition. In this work, we seek to predict the ground-state wavefunction $\Psi_0$, corresponding to the smallest eigenvalue of the Ising Hamiltonian, $E_0$. Wavefunctions are subject to several requirements from the postulates of quantum mechanics \cite{Griffiths2004Introduction}, one of which is that the (Hilbert) vector space of states must possess an inner product $\left(\Psi_1, \Psi_2 \right) \equiv \braket{\Psi_1|\Psi_2}$. Further, the states are normalized such that the overlap of a state with itself is one, $\braket{\Psi_1|\Psi_1}=1$.

The fields $B_x,B_z$ may be used to create a quantum phase transition \cite{McCoy}. When $B_x<1$, the system is in the ferromagnetic phase (spins preferring all up or down in the $z-$basis). As the parameter is swept to $B_x>1$, the system transitions to the paramagnetic phase (either orientation is possible for each spin). Examples are shown in Figure \ref{fig:spin chain}. The magnetic phase of a material has enormous importance in both applied and theoretical condensed matter physics in classification of the electronic properties materials \cite{Ashcroft76}. Further, an understanding of quantum spin forms the backbone of quantum computing \cite{NielsenQuantumComputing}.

\section{Problem Setup}
\label{sec_pgmlsetup}
\noindent
Consider a training dataset $D_{\, Tr} = \{D_{\, L} \! \cup  D_{\, UL}\}$ where $D_L$ corresponds to a labeled dataset obtained by diagonalization solvers and $D_{UL}$ refers to an unlabeled dataset. $D_{\, L}= \{(X^i, \mathbf{H}^i), \ Y^i\}_{i=1}^n$ consists of transverse and longitudinal magnetic field, $B_x$ and $B_z$ respectively, which contribute elements for the corresponding Hamiltonian matrix $\mathbf{H}$. For each data-point, the set of all the spin configurations is denoted by $C = \sigma_{1} \ \sigma_{2} \ ..\ \sigma_{2^N}$ where $\sigma_i$ represents the $i^{th}$ configuration. The ground truth $Y^i$ is the ground-state wavefunction $\Psi^i$ of the Hamiltonian $\mathbf{H}^i$. The set of unlabeled examples, $D_{UL} = \{(X_U^i, \mathbf{H}^i)\}_{i=1}^m$ can be leveraged for unsupervised training. The problem then is to learn an NN model with parameters $\mathbf{\theta}$ to predict the wavefunction $\hat{\Psi}$ of any unseen $\mathbf{H}$, i.e., $\hat{\Psi} = f_{NN}(\mathbf{\theta}; \ X, \mathbf{H})$. 

\begin{figure*}[htb]
	\centering
	\includegraphics[width=0.9\textwidth]{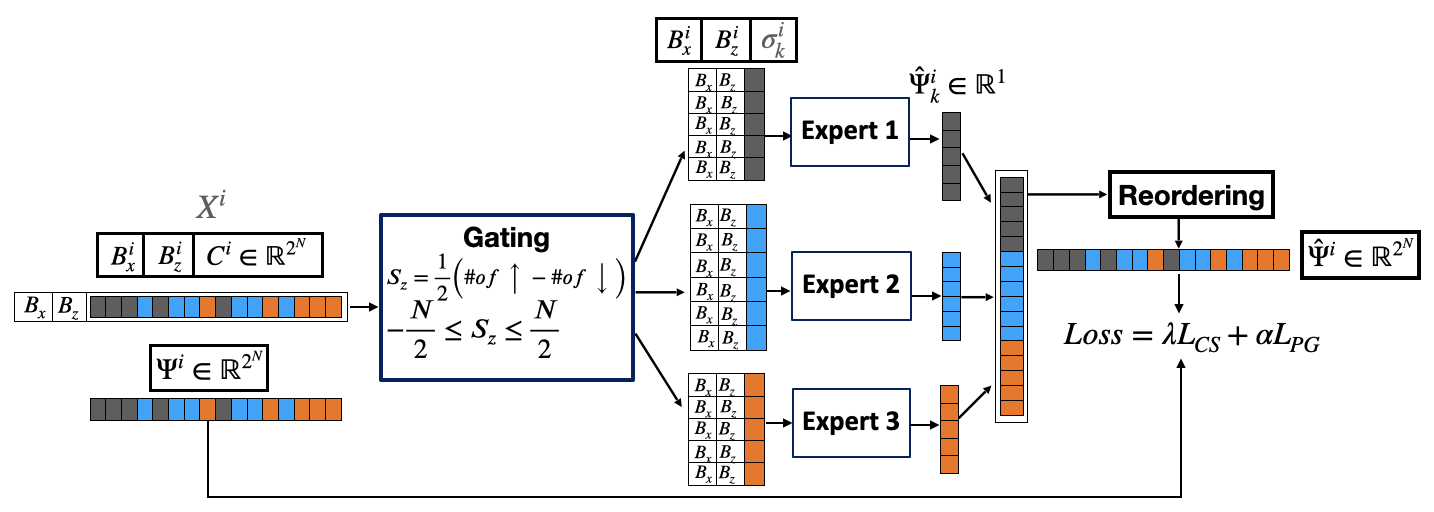}
	\caption{Physics-Guided Mixture-of-Experts (PG-MoE) pipeline. The gating component distributes the $2^N$ configurations and the corresponding coefficients among the experts. The configurations with the same color are mapped to the same expert. Before the loss calculation, the concatenated output from all the experts $\hat\Psi$ are reordered to preserve the configuration ordering of the true $\Psi$. Here, we show $N\!=\!4$ and number of experts as 3.}
	\label{fig:moe}
\end{figure*}
\section{Problem Decomposition}
\label{sec:subspace}
\noindent
We seek to construct a physics-guided decomposition of the input data. Any  decomposition of this nature must satisfy two broad constraints: firstly, it must be physically justifiable, i.e., we may not decompose coupled quantities and, secondly, there must exist a well-defined scientific criterion that can be consistently applied to each input data point. For the Ising model, we choose the total magnetization quantum number, $S_z$ (Equation \ref{eq:sz}), to decompose the input space of $2^N$ configurations into several input sub-spaces. The orthonormality of the configuration basis vectors (Section \ref{sec:quantum}) satisfies the first constraint. As for the second constraint, it is known that spin configurations with comparable values of $S_z$ respond similarly with respect to external physical parameters. Additionally, the value of $S_z$ classifies the magnetic phase of the system, and it can be used to detect the phase transition. These two qualities make $S_z$ a viable choice for determining the decomposition criterion. We therefore use contiguous values of $S_z$ to form the sub-spaces, and learn an ``expert" network for each sub-space. Table \ref{tbl:subspace} illustrates one such decomposition for $N=4$.

\begin{equation}
\label{eq:sz}
  S_z = \frac{(\textnormal{Total \# of} \uparrow - \textnormal{Total \# of} \downarrow)}{2}  
\end{equation}

\begin{table}[htb]
\begin{tabular}{l|l}
\hline
\multicolumn{1}{c|}{\textbf{Sub-spaces}} & \multicolumn{1}{c}{\textbf{configurations}} \\ \hline
\hline
$S_z$ = {-2, -1} & $\{\downarrow \downarrow \downarrow \downarrow\}$, $\{\uparrow \downarrow \downarrow \downarrow\}$, ..., $\{\downarrow \downarrow \downarrow \uparrow\}$ \\
$S_z$ = 0 & $\{\uparrow \uparrow \downarrow \downarrow\}$, $\{\uparrow \downarrow \uparrow \downarrow\}$, ..., $\{\downarrow \downarrow \uparrow \uparrow\}$ \\
$S_z$ = {1, 2} & $\{\uparrow \uparrow \uparrow \uparrow\}$, $\{\downarrow \uparrow \uparrow \uparrow\}$, ..., $\{\uparrow \uparrow \uparrow \downarrow\}$ \\ \hline
\end{tabular}
\caption{One possible decomposition of the configurations based on the $S_z$ value for $N=4$}
\label{tbl:subspace}
\end{table}

For the 10-spin Ising model, we train \textbf{three $\mathcal E$ networks} where the network assignment is determined according to Table \ref{tbl:sz}. 

\begin{table}[htb]
\centering
\begin{tabular}{c|c|c}
\hline
\textbf{Expert No.} & \textbf{$S_z$} & \textbf{\#configurations} \\ \hline \hline
1 & $-5 \leq{S_z} \leq{0}$ & 638 \\
2 & $1 \leq{S_z} \leq{3}$ & 375 \\
3 & $S_z = {4, 5}$ & 11 \\ \hline
\end{tabular}
\caption{A configuration-to-\textit{expert} mapping based on the $S_z$ quantum number}
\label{tbl:sz}
\end{table}

\section{Experimental Plan}
\label{sec:method}

\subsection{Dataset}
We consider an $N\!=\!10$ spin system for the Ising model. The training data $D_{\, Tr}$ comprises 300 sets of physical parameters in the range of $J\!=\!1, \ 0 \! \leq \! B_x \! < \! 1,$ and $0.06 \! < \! B_z \! \leq \! 2$, with 10 and 30 steps for $B_x$ and $B_z$ respectively. The minimum value of $B_z$ is set to 0.06 to eschew ground state degeneracy. From each of the 10 unique bins of $B_x$ values with 30 different $B_z$, we randomly choose 4 data points; 2 for the test set and 2 for the validation set. The resulting test set is denoted by $D_{\, T}$. The distribution of the data is listed in Table \ref{tbl:dataset}. 

Due to the computation overhead of the exact diagonalization to produce annotated data, we also leverage unlabeled data in the training. We test the generalization of the NN for various sizes of annotated and unlabeled data. The network obtains the most supervision when all of the $B_x$ in $D_{\, Tr}$ are used in the annotated set $D_{\, L}$. With the decrease in the amount of data in $D_{\, L}$, the amount of supervision decreases. The most restrictive learning setting occurs when only half of the training data, i.e., from $B_x \! \leq \! 0.5$, make up the annotated set $X$, whereas the other half, from $0.5 \! < \! B_x \! < \! 1$, make up the unlabeled set $X_{\, UL}$.

\begin{table}[htb]
\begin{tabular}{c|c|c|c|c}
\hline
\textbf{$B_x$} & \multicolumn{1}{c|}{\textbf{\begin{tabular}[c]{@{}c@{}}\#Train\\ (labeled)\end{tabular}}} & \multicolumn{1}{c|}{\textbf{\begin{tabular}[c]{@{}c@{}}\#Train\\ (unlabeled)\end{tabular}}} & \multicolumn{1}{c|}{\textbf{\#Valid}} & \textbf{\#Test} \\ \hline \hline
$B_x \! < \! 1$ & 130 & 130 & 20 & 20 \\ 
$1 \! \leq \! B_x \! \leq \! 2$ & 0 & 0 & 0 & 300 \\ \hline
\end{tabular}
\caption{Distribution of train, test and validation data}
\label{tbl:dataset}
\end{table}

Additionally, a dataset $D_{UT}$ is generated from the range $1 \! \leq \! B_x \! \leq \! 2$ to test the generalizability of the trained NN on unseen transverse magnetic field values. Here, $UT$ stands for \textit{Unseen Test}. Specifically, we train the NN with states only in the ferromagnetic phase and hold out the paramagnetic phase states as a stress test.

\subsection{Loss}
\noindent \textbf{Cosine Similarity Loss.} A pure black-box technique to train the NN involves optimizing the network for loss--such as cosine similarity (CS) loss or mean square error (MSE)--between the true wavefunction $\Psi$ and the predicted wavefunction $\hat{\Psi}$. We prefer CS loss over MSE loss because of the normalized inner product requirement satisfied by the wavefunction. Further, the eigenstates may have an overall phase ambiguity \cite{Sakurai:1167961} and with MSE loss the predicted states may not capture phase whereas with CS loss both the magnitude and direction for the ground-state wavefunction will be captured. For $M$ number of data-points, the cosine loss, $L_{\, CS}$, is given by Equation \ref{eq:cosine_loss}. 

\begin{align}
\label{eq:cosine_loss}
L_{\, CS} & = \frac{1}{M}\sum(1 - \cos(\theta))\\
\cos(\theta) & = \frac{\Psi \cdot \hat{\Psi}}{\|\Psi \|\|\hat{\Psi}\|}
\end{align}


\noindent \textbf{Physics Loss.} The physics loss we propose to use is based on the variational principle of quantum mechanics \cite{Griffiths2004Introduction} that states that the expectation value of the energy computed with any state \textit{except} for the ground state will overestimate the ground state energy; $\frac{\bra{\Psi}\mathbf{H}\ket{\Psi}}{\braket{\Psi|\Psi}}\geq E_0$. As such, the energy expectation value may only be a global minimum for the true ground state wavefunction - the quantity the NN learns to predict.
We propose to use the NN predicted wavefunction $\ket{\hat{\Psi}}$ as a variational wavefunction in a new PG loss term such that the network must predict a wavefunction which lowers Equation \ref{eq:pg_loss} as much as possible.

\begin{equation}
\label{eq:pg_loss}
    L_{\, PG}=\frac{1}{M}\sum\frac{\bra{\hat\Psi}\mathbf{H}\ket{\hat\Psi}}{\braket{\hat\Psi|\hat\Psi}}
\end{equation}

\noindent This loss only depends on $\textbf{H}$, which may be computed for any input without re-diagonalizing for the ground truth energy $E_0$.

The overall loss, $L$, is given by the Equation \ref{eq:loss} where $\lambda_{\, CS}$ and $\lambda_{\, PG}$ are hyper-parameters to determine the scaling of each loss.
\begin{equation}
\label{eq:loss}
    L = \lambda_{\, CS} L_{\, CS} + \lambda_{\, PG} L_{\, PG}
\end{equation}

\subsection{Architectures}
\label{sec:arch}
\begin{figure}[htb]
	\centering
	\includegraphics[width=\linewidth]{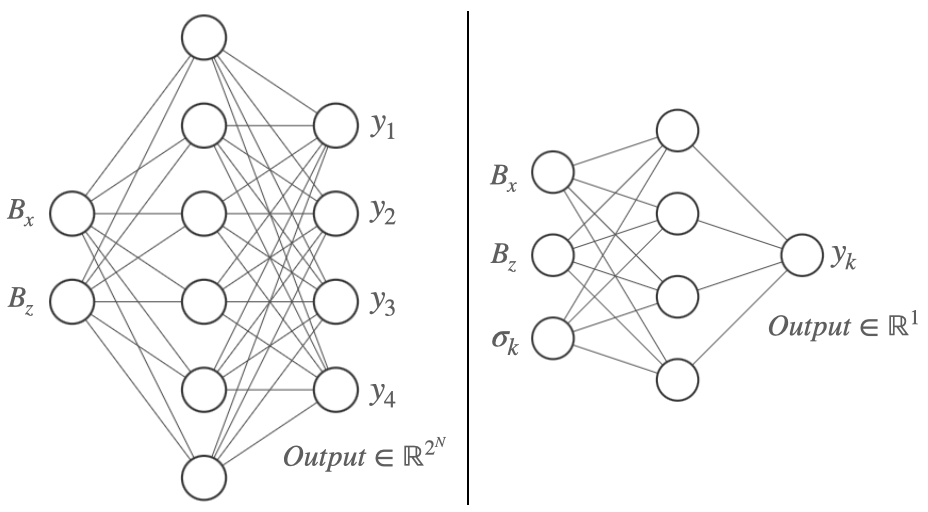}
	\caption{Example architectures for $N\!=\!2$. (Left) The \textit{baseline} model is trained to predict the wavefunction coefficients $\{y_1,\dots,y_{2^N}\}$ for all the configurations at the same time. (Right) The \textit{expert} is tasked to predict only one coefficient $y_k$ corresponding to the configuration $\sigma_k$.}
	\label{fig:experts}
\end{figure}

\noindent \textbf{Baseline: }
Similar to \cite{elhamod2021cophypgnn}, we train a feed-forward network (FFN) with $2^N$ nodes in the output layer to predict the ground-state wavefunction with $2^N$ coefficients. With the $B_x$ and $B_z$ as the inputs, the network is optimized to minimize the loss in Equation \ref{eq:loss}. This serves as our baseline architecture. Figure \ref{fig:experts} illustrates the design of a \textit{baseline} and an \textit{expert} model.\\ 

\noindent \textbf{Physics-Guided Mixture-of-Experts (PG-MoE): }
The PG-MoE consists of a set of \textbf{\textit{$\mathbf{n}$ expert} networks} $\Ex_1, \dots, \Ex_n$, a \textbf{\textit{gating}} component $\mathbf{G}$ and a \textbf{\textit{reordering}} component $\mathbf{R}$. Each expert is a FFN with only 1 output node to predict one coefficient $y_k$ of a wavefunction corresponding to a single configuration $\sigma_k$. Therefore, in addition to $B_x$ and  $B_z$, each expert also takes the configuration $\sigma_k$ as the input. We use a vector of 1s and 0s representing $\uparrow$ and $\downarrow$ to feed $\sigma_k$ to the NN. For example, an input \{1,0,0,0\} represents a configuration $\sigma = \{\uparrow \downarrow \downarrow \downarrow\}$. 

Unlike the gating component used in various MoE architectures in the literature \cite{shazeer2017outrageously} that need a nonlinear optimization routine, our gating component is non-parameterized and does not have to be trained. When given an input with all the $2^N$ configurations, the gating component simply calculates the value of $S_z$ for each and distributes the corresponding batches of $(B_x, B_z, \sigma_k) \rightarrow y_k$ pairs among the experts. The output from all the experts are then concatenated. As shown in Figure \ref{fig:moe}, contiguous configurations do not yield the same value of $S_z$ and thus alter the order of the predicted coefficients in $\hat\Psi$ when concatenated. Therefore, the \textit{reordering} component is used to re-arrange the outputs to their original ordering before calculating the loss. The design of the PG-MoE pipeline is shown in Figure \ref{fig:moe}.

\subsection{Training and Evaluation}
Since the PG loss is independent of the ground-truth $\Psi$, we calculate only the PG loss for data corresponding to unlabeled data $D_{UL}$. For the annotated data $D_L$, the network is optimized for both the $L_{CS}$ and $L_{PG}$. We evaluate the generalization performance of NN model on the test sets $D_T$ and $D_{UT}$ by using mean value of the cosine similarity between our predicted wavefunction $\hat \Psi$ and the ground-truth $\Psi$.

\subsection{Hyperparameter Optimization}
\label{sec:hyperparams}

\subsubsection{Architecture}
We sweep across different combinations of the number of hidden layers ($d$) and number of hidden units ($n$) in each layer to find the architecture that optimizes the validation performance. For the PG-MoE, we use the same architecture for the experts. For the baseline, we use a FFN with $d\!=\!2$ and $n\!=\!2000$. As for each expert, we find $d\!=\!2$ and $n\!=\!200$ to be ideal.
\subsubsection{Training}
\label{sec:hyperparams_train}

Given the small amount of data, we use a small batch size of 8 for all of the training. The loss scaling factors, ($\lambda_{\, CS}$, $\lambda_{\, PG}$), that yield the best validation performance for the baseline and the PG-MoE are (1, 100) and (100, 200) respectively. We train both the baseline and PG-MoE with stochastic gradient descent (SGD) \cite{bottou2018optimization}. We find the training of PG-MoE to be unstable, resulting in the oscillations in the gradient descent, and three ways to mitigate the instability: (1) using a small learning rate, (2) using a momentum \cite{sutskever2013importance} of 0.99, and (3) decaying the learning rate by a factor of 0.987 when the validation loss does not improve for 8 consecutive epochs. For all of the training, we use early stopping \cite{caruana2000overfitting}--training is stopped when the validation performance on the seen data does not improve for 30 epochs. This avoids overfitting of the networks, especially the baseline architecture, which consisting of $\footnotesize\sim$6 million parameters. 

Another key finding is that if PG loss is introduced early in the learning process, the network not only runs the risk of getting stuck in local minima but also the convergence becomes slower. As a result, we only use the cosine loss for the first few epochs before switching to the combination of both the loss components in Equation \ref{eq:loss}. This correspond to $epoch\!=\!80$ and $epoch\!=\!55$ for the baseline and PG-MoE when both are trained for a maximum of 300 and 400 epochs respectively. For each experimental setup, we either run the models for five distinct seeds and choose the one that performs the best across all runs for that study or use mean and standard deviation to combine the five numbers.
\section{Results}
\label{sec:results}

\subsection{Evaluation of Decomposition}
\begin{figure}[h]
\centering
\includegraphics[width=\linewidth]{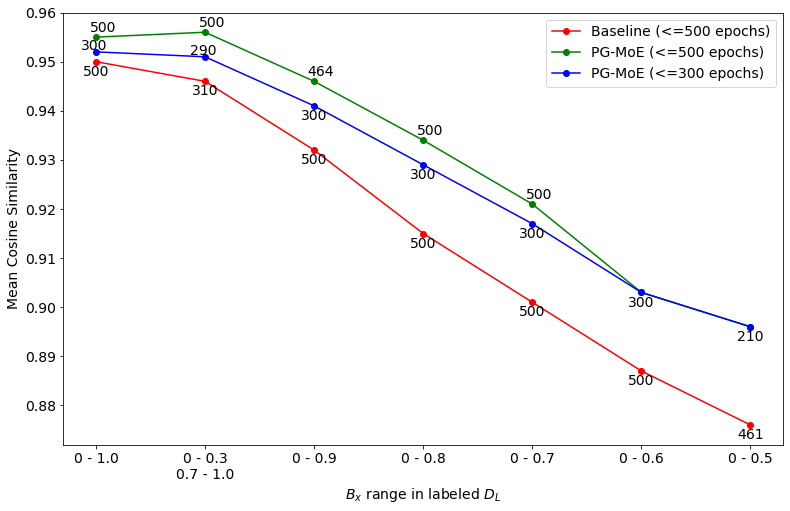}
\caption{Mean cosine similarity on unseen test data $D_{\, UT}$ of the \textit{baseline} and the PG-MoE networks when trained for 300 and 500 epochs. The annotated labels show the epoch where the training early stopped.}
\label{fig:moe_vs_baseline}
\end{figure}

Figure \ref{fig:moe_vs_baseline} compares the generalization performance of the baseline and PG-MoE on the unseen test set $D_{\, UT}$. Both the networks are trained on different subsets of labeled data and optimized for CS loss (i.e., $\lambda_{\, PG} \! = \! 0$). Although PG-MoE outperforms the baseline for all categories of training data, the gap between the two increases as the number of labeled data in $D_{\, L}$ decreases from 260 to 130. For $D_{\, L} \!:B_x \!<\! 0.5$, the mean CS of the PG-MoE on $D_{\, UT}$ is better than baseline by $2\%$ while requiring $2\times$ fewer epochs to train. On average, PG-MoE requires $\sim$60$\%$ fewer epochs than the baseline to achieve similar test performance.

\begin{table*}[]
\begin{tabular}{l|l|c|cr|cr}
\hline
\multicolumn{1}{c|}{\multirow{2}{*}{\textbf{Model}}} & \multicolumn{1}{c|}{\multirow{2}{*}{\textbf{Architecture}}} & \multirow{2}{*}{\textbf{\#params}} & \multicolumn{2}{c|}{\textbf{CS (w/o $L_{PG}$)}} & \multicolumn{2}{c}{\textbf{CS (with $L_{PG}$)}} \\ \cline{4-7} 
\multicolumn{1}{c|}{} & \multicolumn{1}{c|}{} &  & \multicolumn{1}{c|}{$B_x \!< \! 1$} & \multicolumn{1}{c|}{$B_x \!> \! 1$} & \multicolumn{1}{c|}{$B_x \!< \! 1$} & \multicolumn{1}{c}{$B_x \!> \! 1$} \\ \hline
\textbf{Baseline} & $d\_2\_n\_2000$ & \multicolumn{1}{r|}{6093024} & \multicolumn{1}{r}{$0.996\pm0.0008$} & $0.874\pm0.0028$ & \multicolumn{1}{r}{$1.000\pm0.000$} & $0.981\pm0.0003$ \\
\textbf{PG-MoE} & $d\_2\_n\_200$ & \multicolumn{1}{r|}{39903} & \multicolumn{1}{r}{$0.996\pm0.0008$} & ${0.882\pm0.0083}$ & \multicolumn{1}{r}{$1.000\pm0.000$} & ${0.977\pm0.0044}$ \\ \hline
\end{tabular}
\caption{Comparison of cosine similarity of the Baseline and PG-MoE networks on the $D_{\, T}$ ($B_x \!< \! 1$) and $D_{\, UT}$ ($B_x \!> \! 1$) test sets. $d$ and $n$ correspond to the number of hidden layers and the number of neurons in each layer of the NN.}
\label{tbl:pg-moe}
\end{table*}

\subsection{Effect of Physics Loss}
Table \ref{tbl:pg-moe} provides a more detailed comparison of the baseline and the PG-MoE model on both the seen and the unseen test ranges: $D_{\, T}$ and $D_{\, UT}$. To determine how sensitive the networks are, we run both models for 5 different seed values and compute the mean and standard deviation of cosine similarity for each. When trained on $D_{\, L}\!:\!B_x \!< \! 0.5$ and $D_{\, UL}\!: \!0.5 \! < \! B_x \! < \!1$, PG-MoE requires $\sim$150$\times$ lesser parameters than the baseline architecture to achieve competitive performance. For both networks, PG loss yields a perfect cosine similarity on the seen test set $D_{\, T}$. The test performance on $D_{\, UT}$ also improves by $\sim$12$\%$ for both the baseline and the PG-MoE. It is worth noting that the standard deviation in CS for both models is much lower when they are trained with PG loss. This demonstrates how PG loss helps in limiting the convergence region, making the training less sensitive to variations.

Another observation is that the PG-MoE has a larger standard deviation than the baseline for the unseen test $D_{\, UT}$. This is because, as mentioned in Section \ref{sec:hyperparams_train}, the training procedure of PG-MoE is susceptible to oscillations in the gradient descent, leading to variations in performance. Note that the optimization process of training multiple experts in PG-MoE is much more complex than training a single baseline, where the gradient descent of each expert to their respective local minima may show competing directions. Also note that the selection of the loss scaling factors, $\lambda_{\, CS}$ and $\lambda_{\, PG}$ in Equation \ref{eq:loss}, at different stages of PG-MoE training can influence variations in performance. Further algorithmic improvements in training mixture of experts are needed to reduce the variance in PG-MoE results.


\subsection{Annotated Data and Generalization}
\label{sec:sup_gen}

As shown in Table \ref{tbl:physics_loss}, PG loss not only helps generalize better in the unseen test regime, it also helps achieve a consistent test performance as the amount of the labeled data decreases. When the amount of supervised data is decreased by 50\%, i.e., from 260 to 130, PG-MoE optimized only with $L_{CS}$ shows a $\sim$6\% drop in performance whereas the network trained with the PG loss has similar performance.

It is also worth noting that the test performance on unseen $B_x \! > 1$ is not solely related to the number of labeled data-points. Consider two ranges of $D_L$: $B_x \!< 0.3 \cup 0.7 \!< \! B_x \! < \! 1$ and $B_x \!< \!0.6$. Although both have 156 labeled data, the test performance for the former range is better than the latter by $\footnotesize\sim$5\%, when optimized with the black-box loss. This is because the former range includes $B_x$ values close to the ferromagnetic to paramagnetic transition phase (c.f. Section \ref{sec:quantum}), which happens at $B_x \!= \!1$. The inclusion of these data points aids the model's extrapolation for the $D_{UT}$ test set, that primarily consists of data in the paramagnetic phase.


\begin{table}[]
\begin{tabular}{c|cc|cc|cc}
\hline
\multicolumn{1}{l|}{\multirow{2}{*}{\textbf{$\mathbf{|D_L|}$}}} & \multicolumn{2}{c|}{\textbf{$\mathbf{B_x}$ Range}} & \multicolumn{2}{c|}{\textbf{\begin{tabular}[c]{@{}c@{}}CS\\ (w/o $L_{PG}$)\end{tabular}}} & \multicolumn{2}{c}{\textbf{\begin{tabular}[c]{@{}c@{}}CS\\ (with $L_{PG}$)\end{tabular}}} \\ \cline{2-7} 
\multicolumn{1}{l|}{} & \multicolumn{1}{c|}{\textbf{$\mathbf{D_L}$}} & \multicolumn{1}{c|}{\textbf{$\mathbf{D_{UL}}$}} & \multicolumn{1}{c}{\textbf{$\mathbf{B_x} \! < \! 1$}} & \multicolumn{1}{c|}{\textbf{$\mathbf{B_x} \! > \! 1$}} & \multicolumn{1}{c}{\textbf{$\mathbf{B_x} \!< \! 1$}} & \multicolumn{1}{c}{\textbf{$\mathbf{B_x} \! > \! 1$}} \\ 
\hline \hline
260 & \multicolumn{1}{l|}{0 - 1.0} & N/A & 0.998 & 0.952 & 1.000 & 0.976 \\ \hline
156 & \multicolumn{1}{l|}{\begin{tabular}[c]{@{}l@{}}0 - 0.3; \\ 0.7 - 1\end{tabular}} & 0.3 - 0.7 & 0.998 & 0.952 & 1.000 & 0.980 \\ \hline
234 & \multicolumn{1}{l|}{0 - 0.9} & 0.9 - 1 & 0.999 & 0.946 & 1.000 & 0.977 \\ \hline
208 & \multicolumn{1}{l|}{0 - 0.8} & 0.8 - 1 & 0.998 & 0.933 & 1.000 & 0.970 \\ \hline
182 & \multicolumn{1}{l|}{0 - 0.7} & 0.7 - 1 & 0.998 & 0.919 & 1.000 & 0.974 \\ \hline
156 & \multicolumn{1}{l|}{0 - 0.6} & 0.6 - 1 & 0.997 & 0.903 & 1.000 & 0.978 \\ \hline
130 & \multicolumn{1}{l|}{0 - 0.5} & 0.5 - 1 & 0.996 & 0.894 & 1.000 & 0.981 \\ \hline
\end{tabular}
\caption{Cosine similarity for the seen ($B_x \! < \! 1$) and unseen test sets ($B_x \! > \! 1$) for various amounts of labeled data $D_{\, L}$ in the training set. The CS w/o $L_{\, PG}$ column represents PG-MoE optimized with only the black-box loss.}
\label{tbl:physics_loss}
\end{table}

\subsection{Effect of $B_x, B_z$ in Cosine Similarity}
\label{sec:bxbz_cs}

\begin{figure}[htb]
\centering
\includegraphics[width=\linewidth]{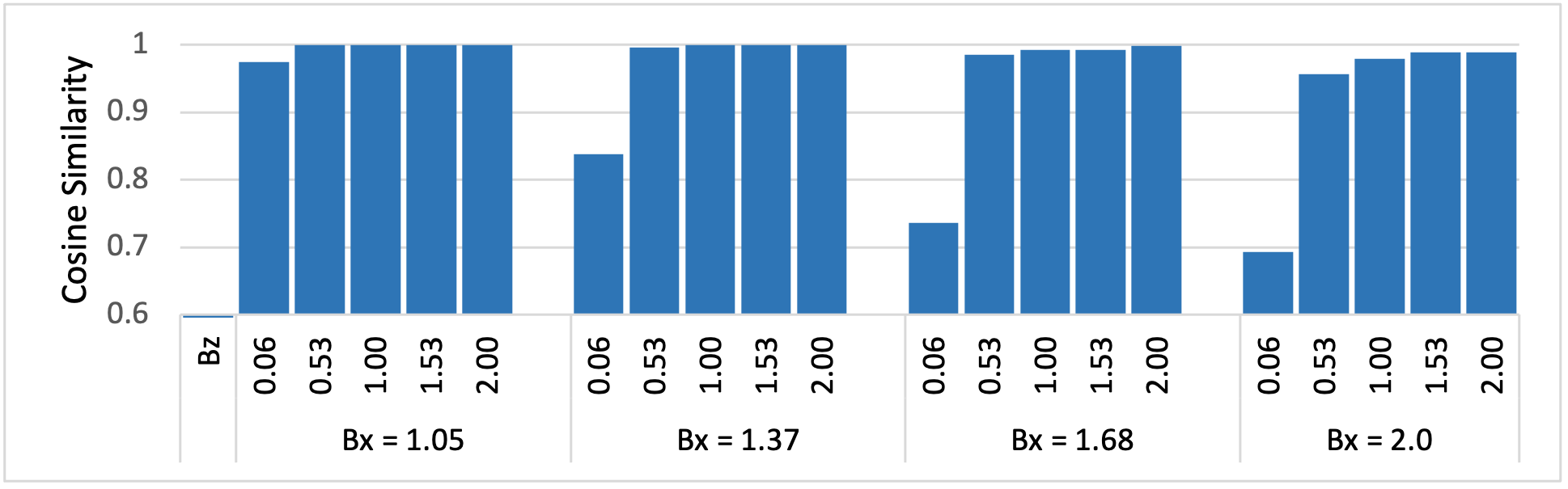}
\caption{Cosine similarity for different $B_x$ and $B_z$ values in the $D_{\, UT}$ test data.}
\label{fig:bxbz_cs}
\end{figure}

The wavefunction prediction is more accurate (in terms of CS) for some ($B_x, B_z$) pairs, whereas for others, it is particularly challenging. The following observations are made from Figure \ref{fig:bxbz_cs}:
\begin{enumerate}
    \item For any given $B_x$, $B_z=0.06$ (lowest value) yields the worst CS, and the performance improves as we increase the value of $B_z$.
    \item If we compare the CS for the same $B_z$, the performance consistently reduces as we increase $B_x$ from 1.05 to 2.0.
\end{enumerate}

It is intriguing that this observation is very consistent with the physical idea of the phase transition. A small but finite value of $B_z$ is enough to align the spins along $z$ direction, and applying a magnetic field along $x$ direction could turn the spin's orientation away from $z$ direction. In other words, $B_z$ and $B_x$ are in fact two competing energy scales to determine the spin orientation. As a result, the stronger $B_x$ is, the fluctuations introduced into the system are larger, which makes the prediction more challenging, and hence the performance is lower.

\begin{table*}[htb]
\begin{tabular}{c|cccc|c}
\hline
\multicolumn{1}{l|}{} & \textbf{$\mathcal E_1$} & \textbf{$\mathcal E_2$} & \textbf{$\mathcal E_3$} & \textbf{$\mathcal E_4$} & \textbf{CS (with $L_{\, PG}$)} \\ \hline \hline
Random & $S_z$ = \{-3, -2, 1, 5\} & $S_z$ = \{-4, 0, 4\} & $S_z$ = \{-5, -1, 2, 3\} & - & ${0.859\pm0.0341}$ \\ \hline
Only-1 & -5 < $S_z$ < 5 & - & - & - & ${0.970\pm0.0083}$ \\ \hline
PG-MoE$_1$ & $S_z$ < 0 & $S_z \! \geq 0$ & - & - & ${0.964\pm0.0167}$ \\
PG-MoE$_2$ & $S_z$ < 0 & $S_z$ = 0 & $S_z$ > 0 & - & ${0.967\pm0.0044}$ \\
PG-MoE$_3$ & $S_z$ <= 0 & $S_z$ = \{1, 2, 3\} & $S_z$ = \{4, 5\} & - & ${\mathbf{0.977\pm0.0044}}$ \\
PG-MoE$_4$ & $S_z$ <= 0 & $S_z$ = \{1, 2, 3\} & $S_z$ = 4 & $S_z$ = 5 & ${0.973\pm0.0031}$ \\ \hline
\end{tabular}
\caption{Mean and standard deviation of cosine similarity on test set $D_{\, UT}$ for various $S_z \! \rightarrow \mathcal{E}_i$ mappings.}
\label{tbl:abl_decom}
\end{table*}
\section{Ablation Studies}
\label{sec:ablation}

We design a number of ablation experiments to explore the impact of various design decisions. 

\subsection{Unlabeled Data and Generalization}
\label{sec:unsup_gen}

\begin{figure}[htb]
\centering
\includegraphics[width=\linewidth]{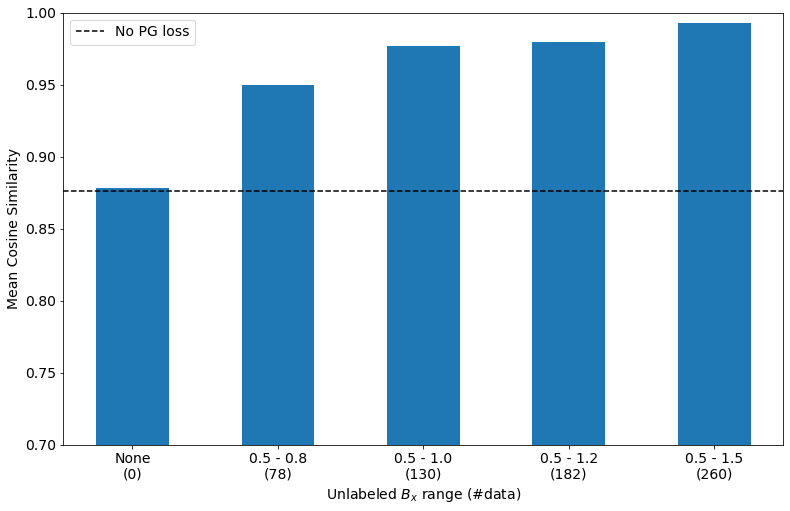}
\caption{Cosine similarity of PG-MoE on unseen test data $1 \!< \! B_x \!< 2$ when trained with annotated data with $B_x \!< \! 0.5$ and different subsets of unlabeled data from $0.5 \!<B_x \!<1.5$. Other than the dashed black line, all the other numbers are from PG-MoE optimized with both CS and PG loss.}
\label{fig:unsup}
\end{figure}

While keeping the labeled data $D_L$ fixed, we vary the amount of unlabeled data in the training from $0.5 \! \leq B_x \!< 1.5$ and test the generalization on the $D_{UT}$ dataset. As shown in Figure \ref{fig:unsup}, PG-MoE trained with PG loss but with no unlabeled data is only slightly better than the one trained only with the black-box loss. The mean CS on the test set consistently increases with the increase in the amount of unlabeled data. The mean CS of the PG-MoE trained with 260 unlabeled data-points is $13\%$ better than the model trained with no unlabeled input.

\subsection{Comparison of Different Decomposition Strategies}
\label{sec:abl_decomp}

In Table \ref{tbl:abl_decom}, we compare the cosine similarity of six types of expert assignments on the unseen test data $D_{UT}$. In addition to the various ways in which we can partition the contiguous values of $S_z$, we also evaluate two comparatively limiting cases of the $S_z \rightarrow \mathcal E$ mapping: (1) randomly grouping $S_z$ in 3 sub-groups for 3 expert architectures, and (2) assigning all the configurations, i.e., all the $S_z$, to only one architecture. For the other partitions, we vary the number of experts from 2 (PG-MoE$_1$) to 4 (PG-MoE$_4$). PG-MoE$_3$ has the same decomposition as in Table \ref{tbl:sz}. For each decomposition, we run the resulting models for 5 distinct seed values and compute the mean and standard deviation of CS for each. 

When optimized with the PG-loss, the random mapping of $S_z \! \rightarrow \! \mathcal E_i$ not only yields the worst test cosine similarity, it also has $\sim$8$\times$ more spread than PG-MoE$_3$. The PG-MoE models become less susceptible to randomization (compare the deviation from PG-MoE$_1$ through PG-MoE$_4$) as $S_z \! \geq \!0$ is sub-divided. We believe this is related to the fact that the phase transition is expected to occur in the range $0 \!\leq S_z \!\leq N/2$, resulting in large variations in the output coefficients. This makes learning in this range more difficult than learning in the $S_z \!\leq \!0$ regime, which comprises of very low coefficient values in the wavefunction. As a result, dividing $S_z \! \geq \! 0$ into multiple partitions helps in decomposing the overall task into simpler and smaller tasks, thus resulting in better CS values of PG-MoE$_2$, PG-MoE$_3$, and PG-MoE$_4$ relative to PG-MoE$_1$. Nevertheless, over-partitioning $S_z \geq 0$ also reduces the training sizes available to each expert and make them under-representative, thus degrading performance. As a result, we can see that the cosine similarity of PG-MoE$_4$ is slightly smaller than that of PG-MoE$_3$.


\subsection{Hyperparameter Selection}

We evaluate the impact of alternate hyperparameter selection with our chosen values in the first row of Table \ref{tbl:design_choices}. Out of five hyperparameters, namely (i) learning rate (\textit{lr}), (ii) epoch where PG-loss is introduced (\textit{pge}), (iii) decay factor for the learning rate scheduler (\textit{$\gamma$}), (iv) momentum (\textit{mom}) for SGD, and (v) loss scaling factors ($\lambda_{\, CS}$, $\lambda_{\, PG}$), we vary one parameter at a time keeping others constant. The table's empty cells imply that the values in the first row are used. 

It is crucial to not train the model with the PG-loss from the very beginning, i.e., \textit{pge = 0}. This is true even for the baseline architecture, and is consistent with the findings in \cite{elhamod2021cophypgnn}. We find PG-MoE architecture to be prone to oscillations, therefore, it skips the global minima in the absence of momentum \textit{(mom = 0)} or learning rate decay \textit{($\gamma \! = \!0$)}, or when the initial learning rate is high ($lr\!=\!0.0004$). While both \textit{mom} and \textit{$\gamma$} has a small search space of \{0.9, 0.985, 0.987, 0.99\}, the parameters ($\lambda_{\, CS}$, $\lambda_{\, PG}$) are particularly difficult to tune. We make two observations while tuning for these scaling factors. First, since the cosine loss can get to a very small value at the later stages of the training, $\lambda_{\, CS}$=1 can lead to vanishing gradients, we, therefore, set this to a high value and empirically find $\lambda_{\, CS}$=100 to be the best choice. (In future work, we will implement adaptive tuning to choose the best scalars). Second, we find that the network is able to extrapolate better when $\lambda_{\, CS} \!< \lambda_{\, PG}$. This can be seen when we compare the scale factor (100, 200) with (100, 100) and (100, 50) in the Table \ref{tbl:design_choices}. 

\begin{table}[htb]
\begin{tabular}{c|c|c|c|c|cc}
\hline
\textbf{lr} & \textbf{pge} & \textbf{$\gamma$} & \textbf{mom} & \textbf{($\mathbf{\lambda_{CS}}$, $\mathbf{\lambda_{PG}}$)} & \textbf{$B_x \!< 1$} & \textbf{$B_x \!> 1$} \\ \hline \hline
0.0003 & 55 & 0.987 & 0.99 & (100, 200) & 1.000 & 0.981 \\ \hline
0.0002 &  &  &  &  & 1.000 & 0.966 \\
0.0004 &  &  &  &  & 0.502 & 0.794 \\ \hline
 & 0 &  &  &  & -1.000 & -0.963 \\
 & 30 &  &  &  & 0.498 & 0.813 \\ \hline
 &  & 0 &  &  & 0.502 & 0.715 \\
 &  & 0.9 &  &  & 1.000 & 0.974 \\
 &  & 0.985 &  &  & 1.000 & 0.979 \\
 \hline
 &  &  & 0 &  & 0.996 & 0.955 \\
 &  &  & 0.9 &  & 1.000 & 0.963 \\ \hline
 &  &  &  & (1,1) & 0.999 & 0.955 \\
 &  &  &  & (100, 50) & 1.000 & 0.954 \\
\multicolumn{1}{l|}{} & \multicolumn{1}{l|}{} & \multicolumn{1}{l|}{} & \multicolumn{1}{l|}{} & (100, 100) & 1.000 & 0.973 \\
\multicolumn{1}{l|}{} & \multicolumn{1}{l|}{} & \multicolumn{1}{l|}{} & \multicolumn{1}{l|}{} & (50, 200) & 1.000 & 0.979 \\ \hline
\end{tabular}
\caption{Effect of changing the hyperparameters in PG-MoE training on the testing performance. The first row corresponds to the choices that yield the best cosine similarity on the test sets when the network is trained with $L_{\, PG}$.}
\label{tbl:design_choices}
\end{table}
\section{Related Work}
\label{sec:related}

\noindent \textbf{Quantum Ising Model. }
Neural networks have been employed to study the quantum Ising model previously. One commonly used method is the restricted Boltzmann machine (RBM) that is designed to learn the distributions of coefficients for spin configurations in the ground state using a generic model. Success has been shown to reproduce the results of physical properties including the magnetization, spin-spin correlation function, and entanglement entropy \cite{rbm0,rbm1,rbm2,iso2018,torlai2016,morningstar2017,gnn2020}. These studies focused on learning particular physical properties instead of the ground state wave function directly. Moreover, the PG loss functions for unlabeled data were not considered.\\


\noindent \textbf{Physics-Guided Machine Learning. }
Physics-guided machine learning is becoming an increasingly common method for solving problems in a wide variety of physics dependent fields such as fluid mechanics \cite{cai2021physicsinformed,ling2016reynolds,raissi2018hidden,wang2020physicsinformed}, electromagnetism \cite{Jin:9474231,Zhang:9350148,Noakoasteen:9158400}, thermodynamic modeling \cite{IHUNDE2022110175,zhang2020thermodynamic,2021thermochemical,patel2020thermodynamically}, and even in medical engineering \cite{10.3389/fphy.2020.00042}. By imposing physical constraints to respect any symmetries \cite{wang2020incorporating}, invariances \cite{ling2016reynolds}, or conservation principles \cite{beucler2019achieving}, researchers are able to constrain the space of admissible solutions to a manageable size even with a few hundred data-points. 

In materials science, most important quantum phenomena are dominated by the ground state solution of the Schr\"odginer equation. Obtaining the ground state solution for a system with many interacting electrons remains one of the most challenging problems in theoretical physics. A traditional approach is to build up a trial function with multiple parameters, namely an {\it ansatz}, to represent the ground state solution, and the parameters in the ansatz are determined by satisfying the physical constraints. Studies of deep learning models using the physics-guided ansatz as the learning target have been performed \cite{2020-dnn-schrodinger,2019-backflow}, but the applicability of the same ansatz to radically different systems is unclear.

The Ising model has also been studied recently in CoPhy-PGNN \cite{elhamod2021cophypgnn}. As is the case in that work, we leverage both unlabeled data and physics loss for better extrapolation. But unlike their neural network, which is also the basis of our baseline architecture, our proposed PG-MoE is scalable for large physical systems. CoPhy-PGNN trains a neural network to predict the entire ground state wavefunction for an $N\!=\!4$ system. With just $2^4=16$ coefficients in the output, it is a much smaller problem as compared to the $N\!=\!10$ system with $2^{10}\!=\!1024$ output values in the wavefunction. With our baseline comparison, we show that PG-MoE can achieve similar test performance as the architecture in CoPhy-PGNN \cite{elhamod2021cophypgnn} while being smaller by more than two orders of magnitude. Additionally, the PG loss presented in CoPhy-PGNN applies to any eigenvector whereas our new loss is true for only the vector we seek to predict.\\

\noindent \textbf{Problem Decomposition. }
Extreme classification tasks in domains like computer vision and NLP have seen a scaling issue similar to Quantum Mechanics. The challenge essentially originates from the necessity to train a deep model to predict millions of classes, which result in a memory and computation explosion in the last output layer. MACH (Merged-Average Classification via Hashing) \cite{medini2019extreme} employs a hashing based divide-and-conquer algorithm to map $K$ classes into $B$ buckets, where $B \! \ll \! K$, and learns a classifier for each bucket. Another method of scaling large neural networks is to use a mixture-of-experts (MoE) \cite{jacobs1991adaptive, eigen2013learning, shazeer2017outrageously}, which involves training multiple expert networks or learners to handle a subset of the entire training set. A gating network is trained to distribute the inputs and combine the corresponding outputs to produce a final output. Although our idea of problem decomposition share the same underlying principle of divide-and-conquer, our approach is different in two respects: (1) we use physics to design the input sub-spaces, and (2) the gating component in our proposed PG-MoE need not be learned and involves a computationally cheap operation (Equation \ref{eq:sz}).

\section{Conclusions and Future Work}
\label{sec:conclusion}

We have demonstrated the use of physics to decompose a complex machine learning problem into smaller problems, which avoids the exponential growth in the neural network's output layer, which is prevalent in high-dimensional eigenvalue problems. The approach leads to learning multiple parameter-efficient expert networks, one for each component problem sub-space, to yield a \textit{Physics-Guided Mixture-of-Experts (PG-MoE)}. The PG-MoE in our case study is $\sim$150$\times$ smaller but has similar extrapolation ability as the model trained on the complex problem. Along with decomposition, we also employed fundamental quantum mechanics in designing a loss function that included the variational method, which better guides the learning of the eigenvector we seek to predict. Notably, while we show the efficacy of our approach for the Ising chain model for 10 spins, our solution is scalable for much larger physical systems. Our work has useful connotations in other scientific applications like quantum spin problems in two-dimensions, interacting fermionic system, and physical systems governed by partial differential equations which possess matrix representation.

For future work, an important topic is to characterize an upper bound for the scalability of the PG-MoE approach, which may be studied by increasing the total number of spins as well as promoting the system to two-dimensions. We will also focus on improving the stability and sensitivity of PG-MoE training. Another intriguing direction is to incorporate traditional theoretical approaches like the quantum Monte Carlo method to train the PG-MoE for large systems. Because labeled data is not easily obtained in these cases, the advantages of PG-MoE for efficient neural network architecture and PG loss for extrapolation would get further tested.

\bibliographystyle{ACM-Reference-Format}
\bibliography{ref}

\appendix

\end{document}